\documentclass{article}
\usepackage{ijcai24}

\usepackage{times}
\usepackage{soul}
\usepackage{url}
\usepackage[hidelinks]{hyperref}
\usepackage[utf8]{inputenc}
\usepackage[small]{caption}
\usepackage{graphicx}
\usepackage{amsmath}
\usepackage{amsthm}
\usepackage{booktabs}
\usepackage{algorithm}
\usepackage{algorithmic}
\usepackage[switch]{lineno}
\usepackage{afterpage}

\title{From RAG to QA-RAG: Integrating Generative AI for Pharmaceutical Regulatory Compliance Process}

\author{
Jaewoong Kim$^1$
\and
Moohong Min$^2$
\affiliations
$^1$Department of Applied Data Science, Sungkyunkwan University \\
$^2$Social Innovation Convergence Program, University College, Sungkyunkwan University
\emails
jwoongkim11@g.skku.edu, iceo@g.skku.edu
}

\begin{document}

\maketitle

\begin{abstract}
    Regulatory compliance in the pharmaceutical industry entails navigating through complex and voluminous guidelines, often requiring significant human resources. To address these challenges, our study introduces a chatbot model that utilizes generative AI and the Retrieval Augmented Generation (RAG) method. This chatbot is designed to search for guideline documents relevant to the user inquiries and provide answers based on the retrieved guidelines. Recognizing the inherent need for high reliability in this domain, we propose the Question and Answer Retrieval Augmented Generation (QA-RAG) model. In comparative experiments, the QA-RAG model demonstrated a significant improvement in accuracy, outperforming all other baselines including conventional RAG methods. This paper details QA-RAG's structure and performance evaluation, emphasizing its potential for the regulatory compliance domain in the pharmaceutical industry and beyond. We have made our work publicly available for further research and development.\footnote{\url{https://github.com/jwoongkim11/QA-RAG}}

\end{abstract}

\section{Introduction}

\subsection{The Advancement of Chatbot}

Recent advancements in Generative AI have significantly enhanced the capabilities of chatbots. The industrial application of these chatbots, powered by Generative AI, is being explored across various sectors \cite{bahrini2023,castelvecchi2023,badini2023}, with the pharmaceutical industry being a notable area of focus.
In the realm of drug discovery, recent study has shown that chatbots, powered by Generative AI, can play a significant role in advancing drug discovery \cite{wang2023,savage2023,bran2023}. Such advancements not only streamline the discovery process but also pave the way for chatbots to suggest novel research ideas or methodologies, enhancing the collaborative aspect of research.
Focusing on healthcare, chatbots are proving to be particularly effective in offering personalized support that can lead to better health outcomes and more effective management of treatments \cite{ogilvie2022,abbasian2023}. These chatbots can provide timely medication reminders, relay information about potential side effects, and even assist in scheduling physician consultations.

\subsection{Needs of Chatbot for Pharmaceutical Regulatory Guidance}
Another crucial domain where Generative AI can be harnessed in the pharmaceutical industry is in ensuring compliance with regulatory guidelines. Navigating the complex and extensive guidelines provided by agencies like the Food and Drug Administration(FDA) and the European Medicines Agency(EMA) is often a daunting and time-consuming task for industry players. The sheer volume of guidelines, combined with their intricate details, can make it challenging for companies to quickly find and apply the relevant information. This often results in increased costs as teams spend valuable time navigating the vast repository of guidelines. A recent study highlighted the financial impact of compliance with regulatory guidelines \cite{crudeli2020calculating}. It revealed that compliance efforts can consume up to 25\% of a medium or large pharmaceutical manufacturing site's operational budget.
In light of these challenges, the pharmaceutical industry requires a more efficient method for navigating and interpreting regulatory guidelines. Large language models (LLMs) can contribute to solving the problem. However, despite their extensive pre-training, LLMs often encounter inherent limitations in accessing knowledge that was not included in their initial training data. Particularly in the realm of pharmaceutical regulatory compliance, a field characterized by its highly specialized and detailed nature, it is clear that such domain-specific knowledge has not been fully included in the training material. As a result, LLMs are likely to be ill-equipped for accurately answering the questions of this field.

The Retrieval-Augmented Generation (RAG) model stands out as a bridge to this gap. It not only utilizes the innate knowledge of these models but also fetches additional information from external sources to generate responses. The RAG framework, as illustrated in the works of \cite{wen2023} and \cite{yang2023}, demonstrates a sophisticated integration of expansive background documents with answers, ensuring comprehensive and accurate responses to queries. These studies highlight the versatility of RAG in diverse applications, from complex story generation to theorem proving. Furthermore, evidence has shown that RAG models excel over typical seq2seq models and certain retrieve-and-extract architectures, particularly in knowledge-dense NLP tasks \cite{lewis2020}.
Despite the advancements in RAG, we recognized that the accuracy of the conventional RAG 
methods may fall short in the regulatory compliance domain, where domain-specific and highly specialized information is required. Hence, we introduce the \textbf{Question and Answer Retrieval Augmented Generation (QA-RAG)}. Tailored for the highly domain-specific sector that needs professional knowledge, the QA-RAG model precisely aligns regulatory guidelines with practical implementation, streamlining compliance in the pharmaceutical industry.

\section{Method}
In this chapter, we will present the QA-RAG model in detail. QA-RAG is a model designed specifically for the highly domain specific areas like pharmaceutical regulatory compliance. The purpose of this approach is to give answers or information to the user’s query related to the guidelines with remarkable accuracy.

\subsection{Overall of QA-RAG Model}

\begin{figure}
    \centering
    \includegraphics[width=1\linewidth]{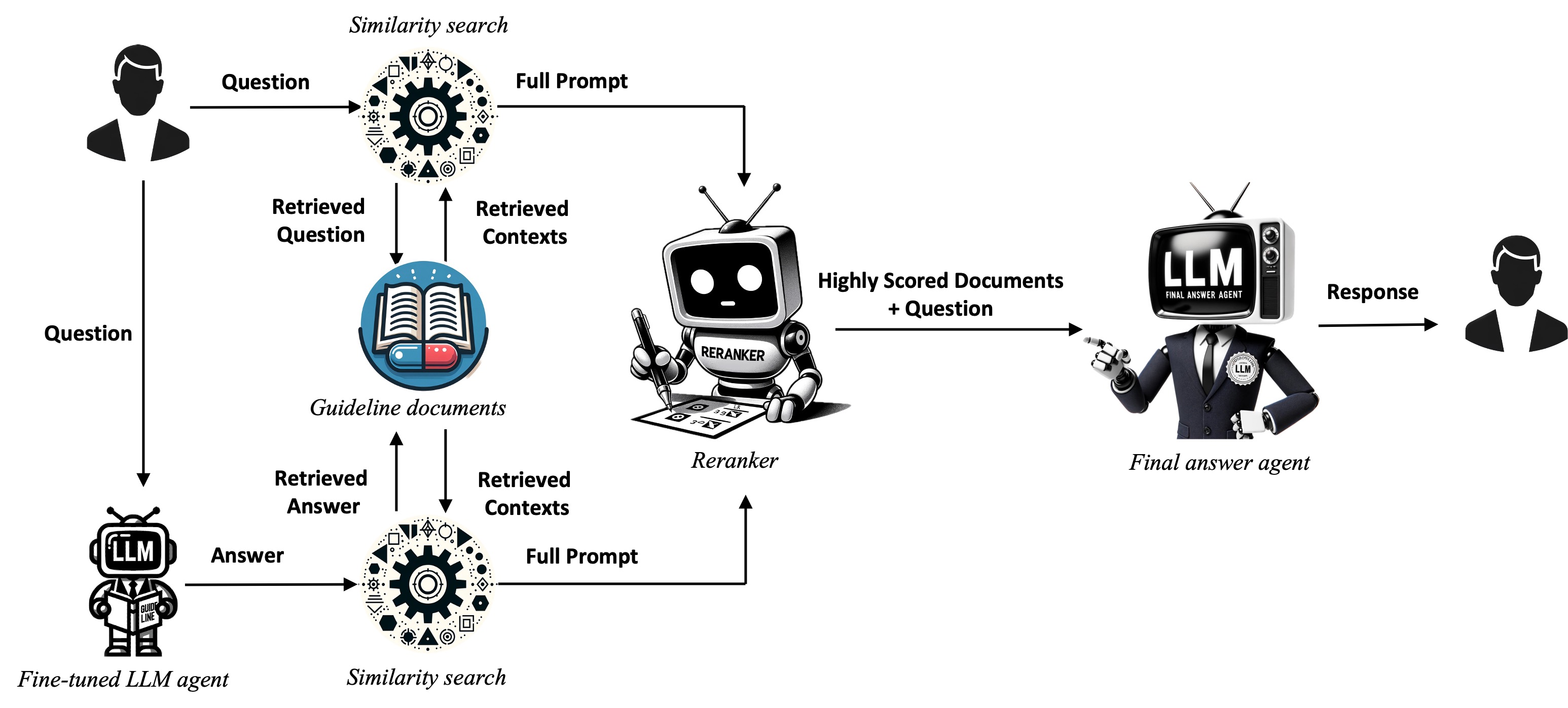}
    \caption{Overall Architecture of QA-RAG}
    \label{fig:QA-RAG_architecture}
\end{figure}

Figure 1 illustrates the overall structure of QA-RAG model. In contrast to the conventional RAG, QA-RAG system utilizes the answer from a fine-tuned LLM agent, with additional support from the query. Half of the documents are sourced through the answer provided by the fine-tuned LLM agent, which are adept at generating contextually rich and accurate responses to the user’s question. The other half of the document set is acquired using the original query. This method not only broadens the scope of the search but also captures a wider array of potentially relevant information.
After obtaining documents through both the answer of the fine-tuned LLM agent and the query, the system then applies a reranking process. This involves evaluating the relevance scores of all retrieved documents with the question and retaining only those with the highest relevance scores. Here’s the breakdown of each part.

\subsection{Document preprocessing \& similarity search}
When it comes to the document retrieval, Sparse retrieval methods such as BM25\cite{robertson2009probabilistic,trotman2014} had been prevalent due to their straightforward approach to matching keywords. However, they can be limited by their inability to capture the deeper semantic meaning of text. Dense retrieval, on the other hand, showed many advantages over sparse retrieval\cite{lee2019,karpukhin2020dense,li2021more}. Dense retrieval approaches go beyond mere keyword matching; they generate vector representations of documents and queries, facilitating the capture of deep semantic meanings. This is crucial in fields requiring high accuracy and contextual understanding, where the relevancy of documents cannot solely be determined by keyword frequency. Given the advantages of dense retrieval, this method was selected for our model.

\subsubsection{Document preprocessing}
We have compiled 1,263 final and valid versions of FDA (Food and Drug Administration) guideline documents regarding the pharmaceutical industry, along with 141 ICH (International Council for Harmonisation of Technical Requirements for Pharmaceuticals for Human Use) guideline documents. This brings the total count to 1,404 documents, each uniquely identified as \(D_i\), where \(i \in \{1, 2, \ldots, 1404\}\).
The FDA is a U.S. federal agency responsible for regulating and overseeing the safety and efficacy of drugs, food, and cosmetics, while the ICH works to harmonize regulatory standards for drug development and registration across different countries, ensuring the safety, quality, and efficacy of medicines.
To effectively extract the content of documents into text, we utilized OCR technology. Specifically, we employed Nougat, a transformer model developed for scientific texts \cite{blecher2023nougat}. This OCR tool is particularly adept at processing technical and scientific documents due to its advanced capabilities. Each document \(D_i\)  processed through OCR is then divided into several chunks, denoted as \(D_{i,j}\), where \(j\) represents the sequence number of each chunk for each document \(i\).
\begin{equation}
D_{i,j}, \text{ where } j \in \{1, 2, \ldots, n\} \text{ for each } i.
\end{equation}
We set the chunk size to 10,000 and the overlap between these chunks to 2,000 characters. The chunk size refers to the maximum number of characters that each chunk of text can contain, and the chunk overlap means the number of characters that overlap between consecutive chunks. The reason we selected a large chunk size and overlap was to obtain a holistic view of overall guideline and to minimize the information loss.

\subsubsection{Document embedding}
For the embedding of the documents, the LLM-Embedder \cite{zhang2023retrieve} model was employed for its demonstrated proficiency in capturing complex semantic relationships within texts. 

\subsubsection{Similarity search}
Through the similarity search, relevant documents are extracted from the database. We employed Facebook AI Similarity Search(FAISS) \cite{johnson2019,danopoulos2019} as the similarity search metric. FAISS is renowned for its efficient and scalable similarity search capabilities, especially in handling large-scale datasets. It offers significant advantages, particularly in terms of speed \cite{mu2019}.

\subsection{Dual-track Retrieval: Leveraging answer of Fine-tuned LLM for Document retrieval.}
We propose a hybrid method that leverages not only the question, but also the hypothetical answer generated by a fine-tuned LLM agent.
In the conventional RAG approach, a single query is employed in similarity search for retrieving relevant documents. However, this method can sometimes be limited in scope, especially when it comes to dealing with the nuances and variability of language. One of the primary challenges is that these methods might miss out on relevant documents due to their dependency on specific keywords or phrases present in the user's query. To address this issue, various solutions have been proposed, including the use of Multiquery retrieval and HyDE\cite{gao2022precise}.

Query transformation for enhanced information retrieval has often been utilized \cite{wang2023query2doc,anand2023query,wang2020deep}. Among such techniques, Multiquery retrieval is an advanced technique that automatically generates multiple queries from the original question with different perspectives. This process, facilitated by a Large Language Model (LLM), retrieves a set of relevant documents for each query, thereby broadening the scope of the search. 

HyDE, on the other hand, leverages hypothetical documents generated in response to the query to enhance the retrieval process. This method involves using an instruction-following language model to generate a text snippet that responds to the query, which is then used in similarity search for document retrieval. The key aspect of this approach is that the generated text doesn't need to be factually accurate, but should capture the relevance pattern of the query, allowing for a more nuanced and effective retrieval of information.

However, Multiquery retrieval is limited as it is still confined to the narrow scope of the user's question, hindering its ability to capture a wide range of information. Also, in domain-specific and highly specialized areas like pharmaceutical regulatory compliance, using a general LLM like what has been done in HyDE often produces very incomplete hypothetical answers, necessitating the employment of a more specialized approach. Recognizing this, we utilized a fine-tuned LLM that has been trained on domain-specific data, which enabled it to generate responses with a level of detail and accuracy akin to that of the expert in the pharmaceutical field. Half of the documents were retrieved using the answers provided by this fine-tuned LLM. To enhance the diversity of search, the other half is sourced using the user’s question. By utilizing both the user's query and the tailored responses generated by the fine-tuned LLM, this dual-track approach achieved a more thorough and nuanced retrieval of information.

\subsubsection{Fine tuning process}

\noindent \textbf{i) Dataset} \\
We used official Q\&A datasets from the FDA for fine-tuning. Due to the comprehensive and sometimes unclear nature of FDA guidelines, a multitude of questions have emerged from both the industry and academia. The FDA offers official responses to these frequently asked questions. We collected them, amounting to 1681 question-answer sets. We designated 85\% of the data for training, 10\% for validation, and the remaining 5\% for testing. The dataset we processed is available online.\footnote{\url{https://huggingface.co/datasets/Jaymax/FDA_Pharmaceuticals_FAQ}}

\noindent \textbf{ii) Base Model} \\
\begin{figure}
    \centering
    \includegraphics[width=1\linewidth]{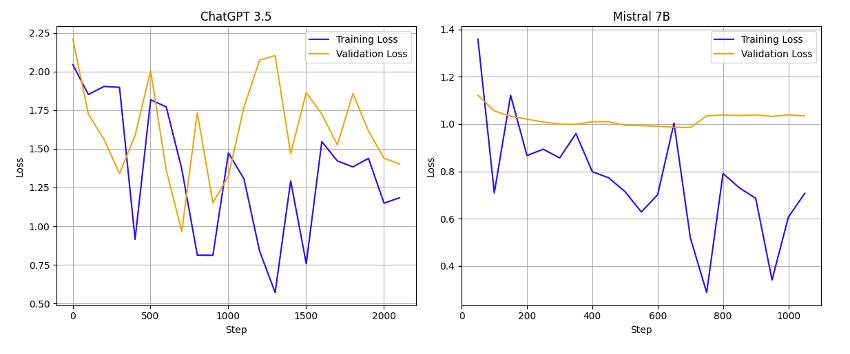}
    \caption{Training and Validation Loss Over Steps}
    \label{fig:Training and Validation Loss Over Steps}
\end{figure}
In this study, we have selected ChatGPT 3.5 - Turbo and Mistral-7B \cite{jiang2023mistral} as our base LLM to be fine-tuned. ChatGPT 3.5 - Turbo model is developed by OpenAI and it boasts the highest performance among those LLMs currently available for fine-tuning, making it a standout choice. The Mistral-7B model, despite having only 7.3 billion parameters, is acclaimed for its high performance. Developed by Mistral AI, this model has demonstrated exceptional performance in various benchmarks, outperforming the Llama 2 13B across all metrics and showing comparable or superior results to Llama 1 34B in key areas such as reasoning, mathematics, and code generation.

For the ChatGPT 3.5 – Turbo model, we conducted fine-tuning over 3 epochs and 2101 steps. As for the Mistral-7B model, to achieve efficient resource handling, we utilized the LoRA (Low-Rank Adaptation) technique \cite{hu2021lora,zeng2023expressive}. LoRA allows for the efficient adaptation of large language models by adjusting a small set of parameters, significantly reducing computational and storage costs. LoRA has been highly successful in modifying large-scale language models \cite{dinh2022lift}. Using LoRA, implemented through the Hugging Face's PEFT library, the Mistral-7B model was fine-tuned over 3 epochs and 1074 steps. Figure 3 shows the result of the fine-tuning process. By the end of the tuning, the loss of both models was significantly reduced, demonstrating the model's enhanced capability to accurately interpret and respond to complex FDA regulatory queries. 

\noindent \textbf{iii) Evaluation} \\
\begin{table}
\centering
\resizebox{\columnwidth}{!}{%
\begin{tabular}{lccc}
\hline
& \textbf{ChatGPT-3.5 Finetuned} & \textbf{Mistral 7B Finetuned} & \textbf{ChatGPT-4} \\
\hline
\textbf{precision} & \textbf{0.579} & 0.485 & 0.505 \\
\textbf{recall} & 0.589 & 0.503 & \textbf{0.622} \\
\textbf{f1} & \textbf{0.578} & 0.489 & 0.555 \\
\hline
\end{tabular}%
}
\caption{Evaluation Results of Fine-Tuned and Standard LLMs on BertScore Metrics}
\label{table:evaluation_results}
\end{table}
We evaluated the performance of the fine-tuned models using BertScore \cite{zhang2019}, a metric for assessing the quality of text generation. BertScore is a sophisticated evaluation method that compares the semantic similarity of machine-generated text to reference text, focusing on precision, recall, and f1 metrics. 

Our comparative analysis included three models: the fine-tuned ChatGPT 3.5 - Turbo, the fine-tuned Mistral-7B, and the non-fine-tuned ChatGPT 4. We added the ChatGPT 4 model for the comparison with the current state-of-art model. The findings revealed a distinct hierarchy in performance. The fine-tuned ChatGPT 3.5 - Turbo model exhibited superior results, leading in both precision and f1 scores. This was followed by the non-fine-tuned ChatGPT 4 model, which, despite not being fine-tuned for this specific domain, still showed commendable capabilities. Lastly, the fine-tuned Mistral-7B model, while effective, ranked third in this comparison.

The fine-tuned ChatGPT 3.5 - Turbo model’s top ranking illustrates its enhanced ability to accurately interpret and respond to complex regulatory queries within the pharmaceutical compliance domain. Hence, for obtaining the best hypothetical answers, we chose to employ the finetuned ChatGPT 3.5-Turbo model as the fine-tuned LLM agent.

\subsection{Reranker}
In sensitive areas such as pharmaceutical regulatory compliance, the accuracy of source documents in answering questions is paramount. Only documents with high relevance to the question should be the basis for the final answer. Furthermore, users may feel the need to compare source documents with the answers to ensure the reliability of the response. So, the pool of retrieved documents should contain only highly relevant documents to the query. However, similarity search methods like FAISS often falter in accurately gauging the contextual relevance due to the limitations in understanding the nuanced interplay between the query and the document content. As a result, while the previous step allowed for the extraction of documents with high relevance to the question, the pool of retrieved documents may also contains less relevant documents.
 
To address this challenge, our initial approach involved adding an extra step featuring a “scoring agent”. This agent was envisioned as a high-performance Large Language Model (LLM) tasked with quantitatively assessing the relevance of each document in the retrieved pool to the query. The core idea was to use this scoring agent to evaluate the contextual alignment of the documents on a ten-point scale. For the LLM to be used, we considered using the ChatGPT-3.5 Turbo model, which can handle the extensive processing demands required for the scoring system.

However, upon further experimentation, we shifted our focus to the reranking approach \cite{nogueira2019,nogueira2020}, specifically employing the BGE reranker \cite{xiao2023cpack}. This decision was based on the reranker’s superior performance in the further experiment described in chapter 3.5, outperforming the custom-developed scoring agent. In this revised system, the BGE reranker evaluates each document in the retrieved documents pool, rank each document’s relevance to the query. 
\begin{equation}
S_{i,j} = \text{Rerank}(Q, D_{i,j}) \quad \text{where} \; j \in \{1, 2, \ldots, N\}
\end{equation}
\begin{equation}
D = \{D_{i,j} \; | \; S_{i,j} \text{ is among the top scores}\}
\end{equation}
Here, \( S_{i,j} \) represents the relevance score assigned to the \( j \)-th chunk of the \( i \)-th document by the reranker, in relation to the query \( Q \). The documents highly ranked constitute the final documents set \( D \). During the final answer phase, responses are generated based on this final documents set.

\subsection{Final Response Generation with Few-Shot Prompting}
To generate the final answer, we introduced the “Final Answer Agent”, for which we employed the ChatGPT-3.5-turbo model.

The final response generation step of the QA-RAG model incorporates a sophisticated few-shot prompting technique \cite{lake2015,miller2000,feifei2006} to enhance the accuracy of the answer. This technique, which outperforms zero-shot inference methods that operate without examples \cite{brown2020}, leverages a composition of an example question-answer set that is fed prior to the question.

Furthermore, to derive an appropriate answer, we designed specific prompts to be fed into the final response agent along with the question and relevant documents. 

\section{Experiment\&Result}

\subsection{Evaluation metric}

For the evaluation, we utilized the LLMs-as-judges metric. Traditional evaluation metrics leveraging n-gram-based evaluation methods like BLEU \cite{papineni2002,reiter2018} have shown limitations in areas outside of Machine Translation \cite{post2018,sulem2018} and ROUGE faces challenges in key factors of machine learning evaluation \cite{grusky2023rogue}, indicating their limitations. Human evaluation has also been a traditional method \cite{awasthi2023}. Since a model is directly judged by humans, it allows for an accurate assessment. However, due to this very reason, human evaluation can be too costly. LLMs-as-judges method could be a good alternative to human evaluation \cite{chiang2023}. This method has shown the highest similarity to human evaluation compared to others \cite{liu2023gpteval,svikhnushina2023}. Furthermore, when utilizing high-performance LLM like GPT-4, the similarity is known to reach up to 80\% \cite{zheng2023judging}.
\subsubsection{Evaluation metric for context retrieval}
Among those LLMs-as-judges metric, we chose the Retrieval Augmented Generation Assessment (Ragas) framework \cite{es2023ragas} for evaluating the accuracy of the context retrieval. Ragas is a framework for the evaluation of the RAG systems. It introduces a suite of metrics for evaluating RAG systems without relying solely on ground truth human annotations. 

The most notable feature of Ragas in evaluating the accuracy of context retrieval is that it does not require a “reference context answer”. Instead, the framework assesses the accuracy of the retrieved context solely based on the question and reference answer. This approach is specialized for situations where no direct reference context is available. The most prominent evaluation metrics for context retrieval Ragas offer include:

\noindent \textbf{i) Context Precision} \\
This assesses the relevance of the retrieved documents to the question. It is especially important when considering the necessity of retrieving guideline documents.

\noindent \textbf{ii) Context Recall} \\
This evaluates the ability of the retrieval system to gather all the necessary information needed to answer the question. It involves using the ground truth answer and an LLM to check if each statement from the answer can be also found in the retrieved context.

Thus, we employed these two metrics to evaluate the quality of the retrieved context in our experiments. By leveraging these metrics, we aimed to provide a comprehensive and objective assessment of the QA-RAG model's performance in retrieving relevant and complete information from the pharmaceutical guidelines.

\subsubsection{Evaluation metric for answer generation}
The final answers generated by the model, based on the retrieved context, could be evaluated by comparing their similarity to the reference answers. For this purpose, we utilized Bertscore \cite{zhang2019}. Given Bertscore’s renowned ability to capture nuanced semantic correlations, it was an ideal choice for comparing the semantic similarity and relevance of the model's responses against the reference answers.

\subsection{Dataset}
The test dataset described in the fine-tuning process section 2.3, which is not used in the training procedure of the fine-tuning, was utilized for the experiments. It consists of answers to frequently asked questions about industry guidelines, compiled directly by the official FDA documents. This dataset's characteristics ensured that it effectively represented the real-world challenges and queries faced in this domain, making it an ideal choice for assessing the performance of the various approaches.

\subsection{Experimental setup}
The evaluation process was designed to compare the performance of the QA-RAG model with various baselines. To ensure a fair evaluation during the experiment, we fixed the number of documents retrieved to 24 for each method, and then narrowed it down to the top 6 documents during the post-processing stage, such as reranking. The accuracy of these contexts was compared across different baselines. Subsequently, the final answer generation step was conducted based on the retrieved context, and the generated answers were also compared across the baselines. The final answer agent used in all cases was the ChatGPT-3.5 Turbo model, and the prompts for answer generation were kept consistent. The experiments include:

\subsubsection{Custom Scoring agent vs. Reranker}
To select an appropriate method for post-processing the retrieved documents, we conducted a comparative analysis between the scoring agent we developed and the reranker method. The scoring agent approach involves feeding the pool of retrieved documents into a Large Language Model (LLM) directly for evaluating relevance with the query. We employed the ChatGPT 3.5 for this purpose. As for the reranker, we opted for the bge-reranker-large \cite{xiao2023cpack}, which is a powerful cross-encoder model. This comparison aimed to evaluate the effectiveness of each method in terms of their ability to accurately prioritize retrieved documents based on their relevance to the query, and then to select the top-ranked ones.

\subsubsection{Evaluation of Context retrieval performance}
The experiment focused on assessing the accuracy of the documents which were selected and finalized through the post-processing stage. The objective was to determine the effectiveness of the QA-RAG compared to various baselines, in accurately extracting relevant documents. The focus was on understanding the precision and relevance of the retrieved documents by each model, to evaluate their overall capability in extracting contexts related to the question from complex pharmaceutical guidelines.

\subsubsection{Evaluation of Answer generation performance}
Following the context retrieval phase, a critical aspect of the evaluation was to assess the QA-RAG model's ability to generate the final answers, in comparison with other baselines. These answers, formulated based on the question and the retrieved contexts, underwent a thorough examination for effectiveness and accuracy.

\subsubsection{Ablation Study}
To further understand the individual contributions of each component in the QA-RAG model, we conducted an ablation study. The QA-RAG setup was configured to retrieve 12 documents based on the question and another 12 from the fine-tuned LLM's answers. Post-processing through the reranking method then narrowed this down to the final 6. We first compared the result of the ``Only hypothetical answer" approach, in which we removed the question-based document retrieval part and retrieved only 12 documents derived from the fine-tuned LLM's answer, again narrowing down to the final 6. Similarly, we compared the ``Only question" approach, which retrieved documents based solely on the question, excluding the fine-tuned LLM's answer.

\subsection{Baseline Selection}

\subsubsection{Question + Hypothetical answer}
This method represents the QA-RAG model, which incorporates both the question and the hypothetical answer derived from the fine-tuned LLM into the retrieval process.

\subsubsection{Multiquery Questions}
We expanded the original question by generating three additional questions, each offering a distinct viewpoint, using the langchain package for implementation \footnote{\url{https://github.com/langchain-ai/langchain}}. We used GPT-4 to generate the additional questions. For each of the four total queries, six contextually pertinent documents were retrieved. After applying the reranker, the top six most relevant documents were extracted.

\subsubsection{HyDE with reranker}
Utilizing a LLM without additional fine-tuning, a single hypothetical document was created by GPT-3.5 Turbo following the ``web search'' prompt described in \cite{gao2022precise}. This document was then used for context retrieval. In contrast to the original HyDE methodology, which used a contriever for post-processing the relevant document pool, we opted for the reranker to maintain consistency with other baselines and ensure fair evaluations. From the hypothetical document, we initially retrieved 24 contexts, out of which the reranker selected the top 6 most relevant ones.

\subsubsection{Only Questions}
Representing the conventional RAG model, this baseline exclusively used the original user question for retrieving documents. This approach was aimed at assessing the impact of solely relying on user queries on the overall performance. From the 24 documents initially retrieved based on the question, the top six were selected based on their relevance using the reranker. 

\subsubsection{Only Hypothetical Answer}
This method relied solely on the fine-tuned LLM's responses to fetch documents, deliberately omitting the use of the original question. This was done to observe the performance changes when only the answer is utilized for document retrieval. Similarly, out of the 24 documents retrieved, only the six most relevant documents were selected after reranking. 

\subsection{Result}

\subsubsection{Reranker vs Scoring agent}

\begin{table}[H]
\centering
\resizebox{\columnwidth}{!}{%
\begin{tabular}{lcccc}
\hline
\textbf{Retrieval metric} & \multicolumn{2}{c}{\textbf{Context\_precision}} & \multicolumn{2}{c}{\textbf{Context\_recall}} \\
\textbf{(Number of document retrieved)} & \textbf{Reranker} & \textbf{Scoring agent} & \textbf{Reranker} & \textbf{Scoring agent} \\
\hline
\textbf{Question(12) + Hypothetical answer(12)} & \textbf{0.717} & 0.454 & \textbf{0.328} & 0.261 \\
\textbf{Multiquery questions(24)} & \textbf{0.564} & 0.36 & 0.269 & \textbf{0.313} \\
\textbf{HyDE with reranker/ScoringLLM (24)} & \textbf{0.673} & 0.43 & 0.283 & \textbf{0.342} \\
\textbf{Only question(24)} & \textbf{0.556} & 0.389 & \textbf{0.27} & 0.263 \\
\textbf{Only hypothetical answer(24)} & \textbf{0.713} & 0.41 & \textbf{0.295} & 0.279 \\
\hline
\end{tabular}%
}
\caption{Comparison results of Reranker vs ScoringLLM}
\label{tab:reranker_scoringllm}
\end{table}

The comparative analysis between the reranker and the scoring agent revealed a consistent superiority of the reranker in terms of context precision and context recall across almost every method, except only for the context recall metric of the Multiquery and HyDE method. This result suggests that although the scoring agent method may have a slight advantage in retrieving relevant information, determined through comparison with the ground truth, the reranker excels in accurately identifying relevant documents in almost every case. Given the reranker's overall superior performance, we selected it as the post-processing method for the QA-RAG model and applied it across all other baseline methods in our experiments.

\subsubsection{Evaluation of Context retrieval performance}

\begin{table}[H]
\centering
\resizebox{\columnwidth}{!}{%
\begin{tabular}{lcc}
\hline
\textbf{Retrieval metric (Number of document retrieved)} & \textbf{Context precision} & \textbf{Context recall} \\
\hline
\textbf{Question(12) + Hypothetical answer(12)} & \textbf{0.717} & \textbf{0.328} \\
\textbf{Multiquery questions(24)} & 0.564 & 0.269 \\
\textbf{HyDE with reranker (24)} & 0.673 & 0.283 \\
\textbf{Only question(24)} & 0.556 & 0.27 \\
\textbf{Only hypothetical answer(24)} & 0.713 & 0.295 \\
\hline
\end{tabular}%
}
\caption{Evaluation of context retrieved based on different retrieval methods.}
\label{tab:context_retrieval}
\end{table}

The QA-RAG model, using a combination of a question and a hypothetical answer from the fine-tuned LLM, achieved the highest context precision (0.717) and context recall (0.328). This superior performance underscores the model's ability to retrieve highly relevant documents. 
In the case of HyDE, it was observed that the performance was surpassed by the ``Only hypothetical answer" approach, where context retrieval was based solely on answers from the fine-tuned LLM. This finding underscores the effectiveness of employing fine-tuned LLM responses, especially in specialized domains. The fine-tuned model's answers, being more aligned with expert knowledge in pharmaceutical regulations, enhance the relevance and accuracy of the retrieved documents.
In contrast, the Multiquery approach, while effective, showed limitations in achieving high precision (0.564) and recall (0.269). This limitation was even more pronounced in the ``Only question" approach, which demonstrated the least effective performance among the methods tested. This highlights the challenge of relying solely on query in areas where domain-specific knowledge is critical.

\subsubsection{Evaluation of Answer Generation performance}

\begin{table}[H]
\centering
\resizebox{\columnwidth}{!}{%
\begin{tabular}{lccc}
\hline
\textbf{Retrieval metric (Number of document retrieved)} & \textbf{precision} & \textbf{recall} & \textbf{f1} \\
\hline
\textbf{Question(12) + Hypothetical answer(12)} & \textbf{0.551} & \textbf{0.645} & \textbf{0.591} \\
\textbf{Multiquery questions(24)} & 0.532 & 0.629 & 0.573 \\
\textbf{HyDE with reranker (24)} & 0.540 & 0.641 & 0.582 \\
\textbf{Only question(24)} & 0.540 & 0.636 & 0.581 \\
\textbf{Only hypothetical answer(24)} & 0.539 & 0.642 & 0.583 \\
\hline
\end{tabular}%
}
\caption{Evaluation of generated final answer synthesized with different retrieval methods.}
\label{tab:final_answer_evaluation}
\end{table}

The evaluation of final answer generation indicates similar findings. The QA-RAG model achieved the highest scores in precision (0.551), recall (0.645), and F1 (0.591). Notably, the F1 score, a metric that combines precision and recall, exactly matched the top 3 rankings in context retrieval performance, demonstrating the efficacy of employing high-accuracy contexts in generating precise responses.

\subsubsection{Ablation study}

\begin{table}[H]
\centering
\resizebox{\columnwidth}{!}{%
\begin{tabular}{lccc}
\hline
\textbf{Retrieval metric (Number of document retrieved)} & \textbf{Context precision} & \textbf{Context recall} \\
\hline
\textbf{Question(12) + Hypothetical answer(12)} & 0.\textbf{717} & \textbf{0.328} \\
\textbf{Only question(12)} & 0.559 & 0.308 \\
\textbf{Only hypothetical answer(12)} & 0.700 & 0.259 \\
\hline
\end{tabular}%
}
\caption{Ablation study results of QA-RAG model.}
\label{tab:ablation_study}
\end{table}

The ablation study provided valuable insights into the distinct components of the QA-RAG model. Focusing on the hypothetical answer component alone, the model achieved an impressive context precision of 0.700, lower by just 0.017 points than the full model's performance. Conversely, removing the hypothetical answer element and relying solely on the user's question led to a marked drop in context precision to 0.559. In terms of context recall, the ``Only question" approach achieved a slightly higher score of 0.308 compared to the ``Only hypothetical answer" method (0.259). The difference in context precision scores between the ``Only question" (0.559) and ``Only hypothetical answer" (0.700) – more pronounced than in context recall – highlights the crucial role that hypothetical answers play in enhancing precision, suggesting their significant contribution to the model's overall accuracy.

\section{Conclusion}

\subsection{Summary of Findings}

Our investigation into the QA-RAG model within the regulatory compliance domain reveals its effectiveness in merging generative AI and RAG with pharmaceutical regulatory guidelines. The model provides accurate and contextually relevant document retrieval, ultimately delivering precise answers. This is especially crucial in the pharmaceutical sector, where adherence to regulatory standards is critical.
Key findings from our research include:

\subsubsection{Superior Performance Driven by Utilization of Answers}

In our experiments, strategies that incorporated answers for document retrieval exhibited notable advantages. The QA-RAG model, employing a hypothetical answer from a fine-tuned LLM, achieved the highest context precision and recall score. Following closely was the ``Only hypothetical answer" approach, which exclusively used a fine-tuned LLM-generated answer and secured the second-highest context precision and recall score. Furthermore, HyDE, which utilized an answer derived from a general LLM that is not fine-tuned, also achieved the third-highest ranking. It emphasizes the advantage of answer-based retrieval strategies in document precision.

\subsubsection{Advantages of Hybrid Query-Answer Approach}

The ablation study results underlined the importance of a balanced hybrid question-answer approach in the QA-RAG model. While the hypothetical answer component was vital for high precision and recall, integrating the original question also enhanced the model’s overall performance. By effectively merging these two elements, the QA-RAG model optimizes its retrieval accuracy and relevance, proving the value of this combined approach.

\subsubsection{Impact of Fine-Tuned LLM in Retrieval}

The significance of the fine-tuned LLM in the QA-RAG model is validated by its strong performance in our tests. In the context retrieval experiment, the approaches using the fine-tuned LLM (``Only hypothetical answer" and ``Question + Hypothetical answer") ranked among the top two in context precision and recall. Similarly, in the answer generation evaluation, these two approaches again secured the top positions in f1 scoring. This consistent high ranking across different metrics underscores the fine-tuned LLM's critical role in extracting pertinent documents. By providing accurate answers tailored to pharmaceutical regulations, it effectively guides the retrieval of relevant documents.

\subsection{Implications for the Pharmaceutical Industry}
The successful integration of the QA-RAG model into the pharmaceutical industry’s regulatory compliance domain can have following implications:

\subsubsection{Streamlining Regulatory Compliance}
The QA-RAG model with pharmaceutical regulatory guidelines streamlines the compliance process by efficiently providing information through Q\&A. This not only reduces the time and resources required for navigating complex regulations but also facilitates more informed decision-making.

\subsubsection{Reduction in Dependency on Human Expertise}
The model reduces reliance on extensive human expertise traditionally required in this field. By automating parts of the compliance process, it allows experts to focus on more strategic tasks, thereby optimizing the overall workflow.

\subsubsection{Pioneering the Use of Generative AI in Pharmaceutical Regulatory Compliance Domain}
As one of the first instances of employing generative AI within the realm of pharmaceutical regulatory compliance, the QA-RAG model sets a precedent. It illustrates the effective strategy for applying generative AI and RAG in pharmaceutical regulatory compliance, providing a cornerstone for future research.

\subsection{Final Thoughts}

In conclusion, the QA-RAG model marks a step forward in the application of generative AI in pharmaceutical regulatory compliance. It stands out as one of the first models to leverage high-performance Large Language Models (LLMs) for navigating the complex landscape of regulatory guidelines in the pharmaceutical industry. Its enhanced capabilities in document retrieval and answer generation establish it as a more suitable approach compared to the conventional RAG.

Moreover, the adaptable design of the QA-RAG model shows potential for use in other industries that deal with highly domain specific information and require professional analysis. Sectors such as legal compliance, financial regulation, and academic research could greatly benefit from the model's advanced capabilities. Its application could revolutionize the way organizations across various industries manage large data, leading to swifter and more accurate information retrieval that enhances decision-making.

However, like any emerging technology, the long-term implications of the model within various industries will require ongoing evaluation and refinement. The integration of generative AI in highly specialized fields will raise questions about the model's adaptability to nuanced changes in data and industry practices. Thus, future developments should focus on proving the model's sustained effectiveness, ensuring it remains a robust tool in the face of ever-changing landscapes. Furthermore, it’s crucial to keep enhancing the model’s performance by staying aligned with the evolving generative AI technologies.

\section*{Ethical Statement}

In the development and application of the QA-RAG model, we emphasize its role as a complementary tool for professionals in the pharmaceutical field. While the model  enhances the efficiency and accuracy of navigating complex guidelines, it is designed to augment, not replace, human expertise and judgment.

The dataset used for training and evaluating the model consists of publicly accessible documents from the Food and Drug Administration (FDA) and the International Council for Harmonisation of Technical Requirements for Pharmaceuticals for Human Use (ICH), adhering to all applicable data privacy and security protocols. 

\section*{Acknowledgments}

This work was supported by the National Research Foundation of Korea(NRF) grant funded
by the Korea government(MSIT) (No. RS-2022-00166729).

We acknowledge the use of ChatGPT, developed by OpenAI, for generating figures used in this paper to illustrate the model's design and functionality.

\bibliographystyle{named}
\bibliography{QA-RAG}

\end{document}